\documentclass[10pt,journal,compsoc]{IEEEtran}
\usepackage{amsmath}
\usepackage{array}
\usepackage{amsthm}
\usepackage{mdwmath}
\usepackage{color}
\usepackage[justification=centering]{caption}
\ifCLASSOPTIONcompsoc
  \usepackage[nocompress]{cite}
\else
  \usepackage{cite}
\fi

\ifCLASSINFOpdf
   \usepackage[pdftex]{graphicx}
  \graphicspath{{../pdf/}{../jpeg/}}
  \DeclareGraphicsExtensions{.pdf,.jpeg,.png}
\else
  \usepackage[dvips]{graphicx}
  \graphicspath{{../eps/}}
  \DeclareGraphicsExtensions{.eps}
\fi

\begin{document}
\title{Coverless information hiding based on Generative Model}

\author{Xintao Duan, Haoxian Song% <-this % stops a space
\IEEEcompsocitemizethanks{\IEEEcompsocthanksitem Xintao Duan, Haoxian Song were with School of Computer and Information Engineering, Henan Normal University, Xinxiang, Henan,
CHINA, 453007.\protect\\
E-mail: duanxintao@126.com
}% <-this % stops an unwanted space
\thanks%{Manuscript received may 30, 2017; revised May 30, 2017.}
}

%\markboth{Journal of JOURNAL OF IEEE ACCESS,~Vol.~14, No.~8, August~2015}%
%{Shell \MakeLowercase{\textit{et al.}}: The capacity of Device-to-device communication underlaying cellular networks with two-hop links}

\IEEEtitleabstractindextext{%
\begin{abstract}
A new coverless image information hiding method based on generative model is proposed, we feed the secret image to the generative model database, and generate a meaning-normal and independent image different from the secret image, then, the generated image is transmitted to the receiver and is fed to the generative model database to generate another image visually the same as the secret image. So we only need to transmit the meaning-normal image which is not related to the secret image, and we can achieve the same effect as the transmission of the secret image. This is the first time to propose the coverless image information hiding method based on generative model, compared with the traditional image steganography, the transmitted image does not embed any information of the secret image in this method, therefore, can effectively resist steganalysis tools. Experimental results show that our method has high capacity, safety and reliability.
\end{abstract}

}

\maketitle

\IEEEpeerreviewmaketitle

\IEEEraisesectionheading{\section{Introduction}\label{sec:introduction}}

\IEEEPARstart{M}{ost} of the current information hiding technologies embed the secret information into the carrier with the slight modification of the carrier data (digital image, video and audio), and hide the dense carrier as a disguise of the secret information. The popularization of personal computers and the proliferation of multimedia data on the internet have provided convenient conditions for implementing information hiding and have made information hiding rapid development. However, the pace of development in recent years has slowed down. The main reason is that at the same time as the development of information hiding, the detection technology for hidden information, also called steganalysis, has also been rapidly developed. The technology is based on the statistical anomaly of the carrier data caused by information embedding to determine whether the secret information exists, and it has posed a serious threat to information hiding. According to the different hiding methods, the commonly used steganography methods are divided into two types: space domain hiding method and transform domain hiding method. The space domain hiding method has the adaptive LSB hiding method\cite{yang2008adaptive}, the spatial adaptive steganography algorithm S-UNIWARD \cite{Holub2014Universal}, HUGO\cite{Pevn2010Using}, WOW\cite{Holub2012Designing} and so on. The transform domain method is to modify the host image data to change some statistical features to achieve data hiding, such as the hidden method in DFT(discrete Fourier transform) domain\cite{Ruanaidh1996Phase}, DCT (discrete cosine transform) domain\cite{Cox1999Secure}, and DWT (discrete wavelet transform) domain\cite{Lin2008An}. These methods inevitably leave some modifications to the carrier. In order to fundamentally resist the detection of various detection algorithms, this paper presents a new coverless image information hiding method based on generative model. As shown in Fig.1, we only need to deliver a meaning-normal image which is not related to the secret image to the receiver, so that the receiver can generate an image visually the same as the secret image without worrying about the analysis of the steganography, even less attack.
\begin{figure}[ht]
\centering
\includegraphics[width=3.5in]{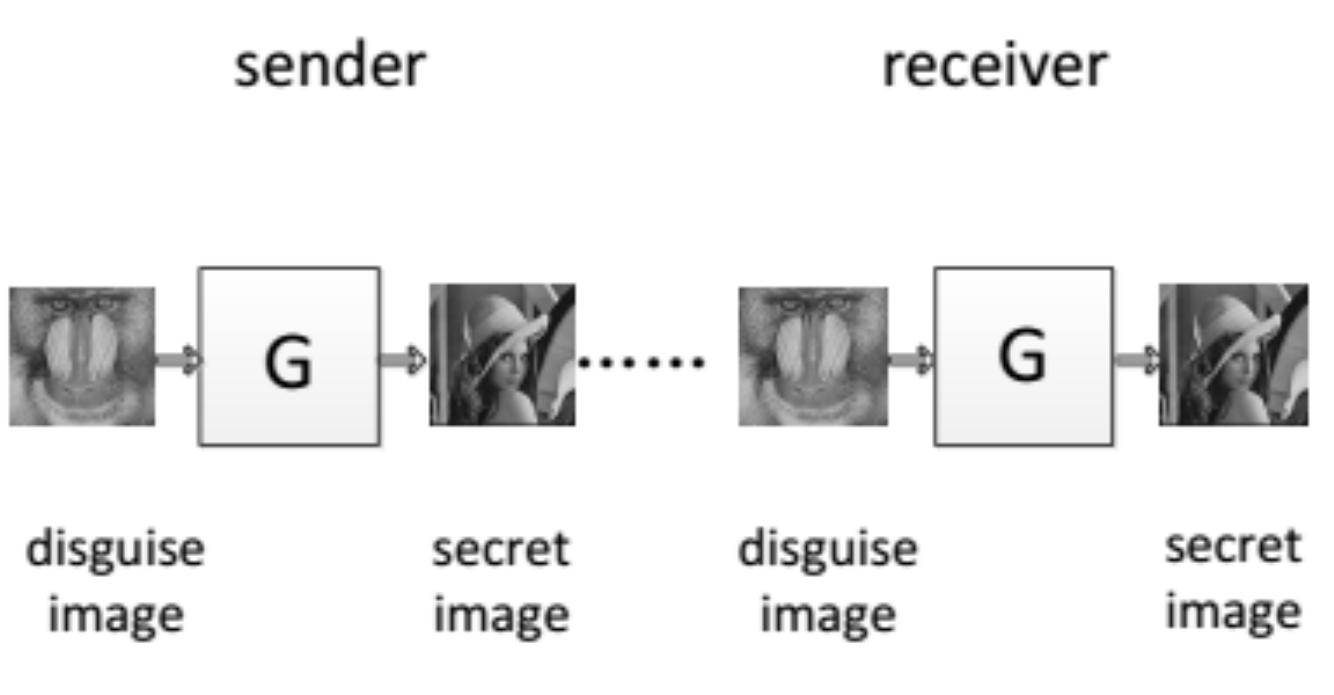}
% where an .eps filename suffix will be assumed under latex,
% and a .pdf suffix will be assumed for pdflatex; or what has been declared
% via \DeclareGraphicsExtensions.
\caption{}
\label{fig_1}
\end{figure}

As mentioned above, we proposed a new approach to hide image information, it can generate visually the same image as the secret image by sending a generated image that is not related to the secret image. The transmitted image is only a normal-meaningful image rather than the image embedded any information of the secret image, and also achieve the same effect as transferring the secret image. This method can effectively resist steganalysis tools, and greatly improves the security of the image. To summarize, the major contributions of our work as below:

\begin{itemize}
\item{强调}We do not need to pass the secret image, on the contrary, we transmit a meaning-normal image which is completely unrelated to the secret image. This method has high security.
\end{itemize}

\begin{itemize}
\item{强调}The image we transmit does not embed any secret information, it is a normal image, and the image steganography analysis does not work.
\end{itemize}

\begin{itemize}
\item{强调}As long as the training is enough, this effect can be achieved and the capacity is large.
\end{itemize}

\section{Related work}
Restricted Boltzmann Machines (RBMs) \cite{Smolensky1986Information}, deep Boltzmann machines (DBMs) \cite{Salakhutdinov2009Deep} and their numerous variants are undirected graphical models with latent variables. The interactions within such models are represented as the product of unnormalized potential functions, normalized by a global summation/integration over all states of the random variables. This quantity and its gradient are intractable for all but the most trivial instances, although they can be estimated by Markov chain Monte Carlo (MCMC) methods. Mixing poses a significant problem for learning algorithms that rely on MCMC \cite{Bengio2012Better},\cite{Bengio2014Deep}. Deep belief networks (DBNs) \cite{Hinton2006A} are hybrid models containing a single undirected layer and several directed layers. While a fast approximate layer-wise training criterion exists, DBNs incur the computational difficulties associated with both undirected and directed models. Variational Auto-Encoders (VAEs) \cite{glorot2011deep} and Generative Adversarial Networks (GANs) \cite{Bengio2013Generalized} are well known to us. VAEs focus on the approximate likelihood of the examples, and they share the limitation of the standard models and need to fiddle with additional noise terms. Ian Goodfellow put forward GAN \cite{Goodfellow2014Generative} in 2014. Goodfellow theoretically proved the convergence of the algorithm, and when the model converges, the generated data has the same distribution as the real data. GAN provides a new training idea for many generative models and has hastened many subsequent works. GAN takes a random variable (it can be Gauss distribution, or uniform distribution between 0 and 1) to carry on inverse transformation sampling of the probability distribution through the parameterized probability generative model (it is usually parameterized by a neural network model), then a generative probability distribution is obtained. The GAN model includes a generative model G and a discriminative model D. The training objective of the discriminative model D is to maximize the accuracy of its own discriminator, and the training objective of generative model G is to minimize the discriminator accuracy of the discriminative model D. The objective function of GAN is a zero-sum game between D and G and also a minimum - maximization problem. GAN adopts a very direct way of alternate optimization, and it can be divided into two stages. In the first stage, the discriminative model D is fixed, the generative model G is optimized to minimize the accuracy of the discriminative model. In the second stage, the generative model G is the fixed in order to improve the accuracy of the discriminative model D. As a generative model, GAN is directly applied to modeling of the real data distribution, including generating images, videos, music and natural sentences, etc. Because of the mechanism of internal confrontation training, GAN can solve the problem of insufficient data in some traditional machine learning. GANs offer much more flexibility in the definition of the objective function, including Jensen-Shannon, and all f-divergences \cite{Hinton2012Improving} as well as some exotic combinations. Therefore, it can be used in semi-supervised learning, unsupervised learning, multi-view learning and multi-tasking learning. In addition, it has been successfully used in reinforcement learning to improve its learning efficiency. Although GAN is applied widely, there are some problems with GAN, difficulty in training, lack of diversity. Besides, generator and discriminator cannot indicate the training process. On the other hand, training GANs is well known for being delicate and unstable. The better discriminator is trained, the more serious gradient of the generator disappears, leading to gradient instability and insufficient diversity. WGAN (Wasserstein Generative Adversarial Networks \cite{Arjovsky2017Towards}, \cite{Arjovsky2017Wasserstein}) is an improvement to GAN, and it applies Wasserstein distance instead of JS divergence in the GAN. Compared to KL divergence and JS divergence, the advantage of Wasserstein distance is that it can still reflect their distance even if there is no overlap between the two distributions. At the same time, the problem of training stability and process indicating are solved.

Therefore, this paper chooses Wasserstein GAN so as to guarantee training stability instead of GAN. It is no longer necessary to carefully balance the training extent between generator and discriminator. It basically solves the problem of collapse mode and ensures the diversity of samples.

\section{Approach}

\begin{figure}[ht]
\centering
\includegraphics[width=3.8in]{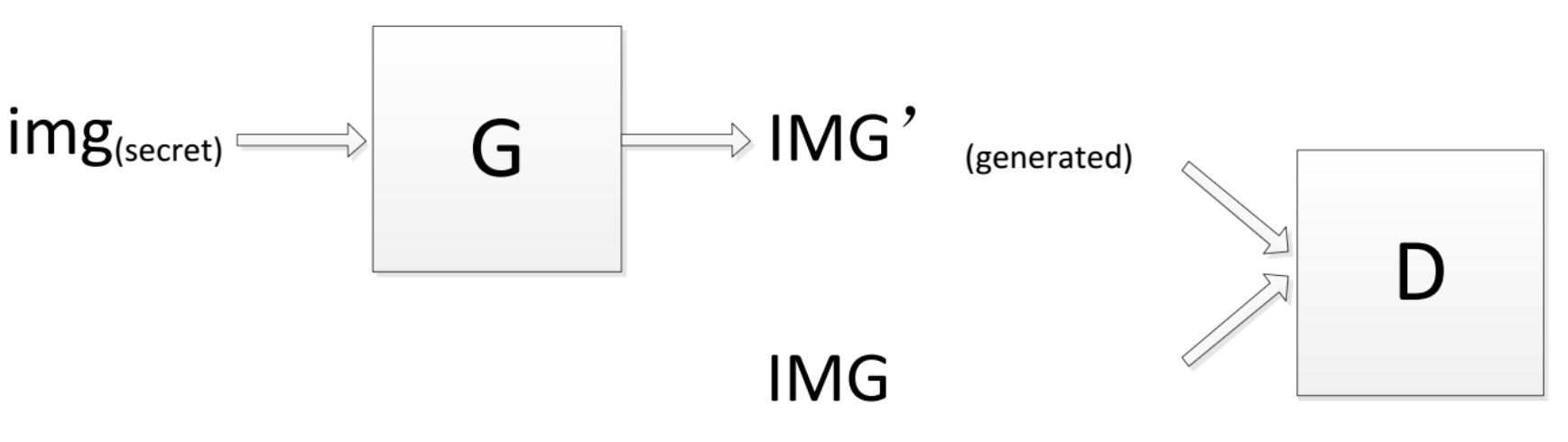}
% where an .eps filename suffix will be assumed under latex,
% and a .pdf suffix will be assumed for pdflatex; or what has been declared
% via \DeclareGraphicsExtensions.
\caption{}
\label{fig_2}
\end{figure}

\begin{figure}[ht]
\centering
\includegraphics[width=3.5in]{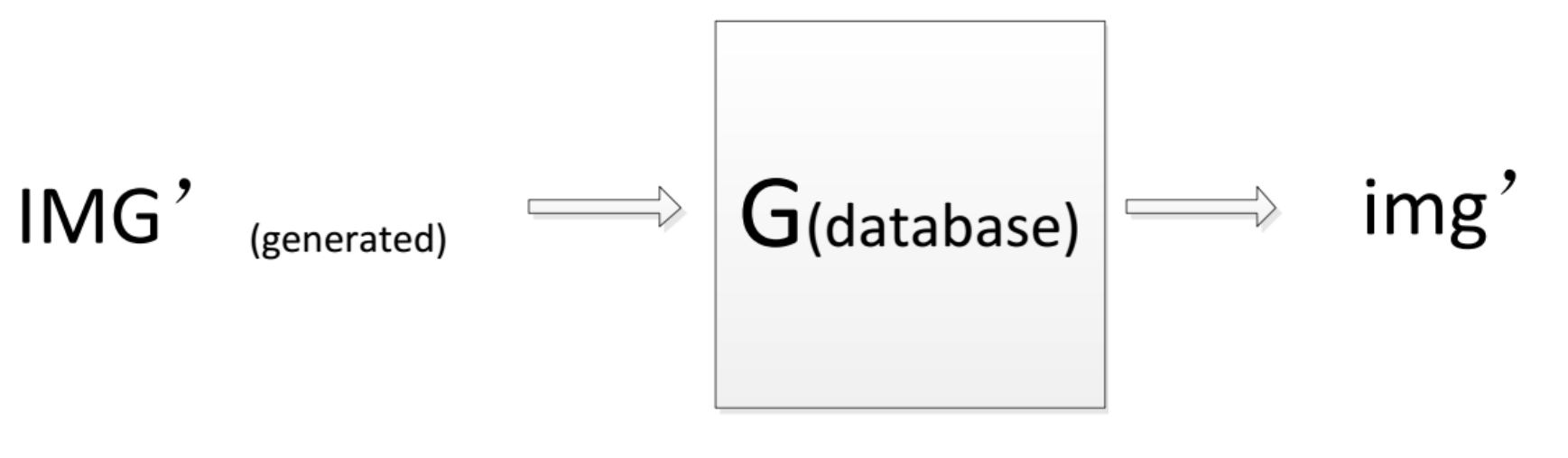}
% where an .eps filename suffix will be assumed under latex,
% and a .pdf suffix will be assumed for pdflatex; or what has been declared
% via \DeclareGraphicsExtensions.
\caption{}
\label{fig_3}
\end{figure}

The WGAN model is applied to generate the handwritten word by feeding the random noise z, but when the random noise z is changed to a secret image img, the model can still generate the meaning-normal and independent image IMG' which is not related to the secret image we want to transmit. These several images taken from the standard set of images were evaluated in the paper, they are Lena, baboon, cameraman and peppers, and they have the same size as 256 by 256. The feed is the secret image, and we train the generative model database through the WGAN, then it can generate a meaning-normal and independent image which is not related to the secret image. So we transmit the meaning-normal image to the receiver, and this generated image is fed to the generative model database to generate another generated image visually the same as the secret image. The flow charts of the whole experiment are shown in Fig.~\ref{fig_2} and Fig.~\ref{fig_3}.


D and G play the following two-player minimax game with value function \[V(G;D)\] in WGAN:

\[\mathop {\min }\limits_G \mathop {\max }\limits_D V(D,G) = {E_{x \sim pdata(x)}}[logD(x)] +
{E_{z \sim {p_z}(z)}}[log(1 - D(G(z)))]\]

D(x) represents the probability that x came from the data rather than pg. We train D to maximize the probability of assigning the correct label to both training examples and samples from G. We simultaneously train G to minimize  \[log(1 - D(G(z)))\].
We first make a gradient ascent step on D and then a gradient descent step on G, then the update rules are:

Keeping the G fixed, update the model D by\[{\theta _D} \leftarrow {\theta _D} + {\gamma _D}{\nabla _D}L\]with

\[{\nabla _D}L = \frac{\partial }{{\partial {\theta _D}}}\{ {E_{x \sim Pdata(x)}}[logD(x,{\theta _D})] +
{E_{z \sim Pnoise(z)}}[log(1 - D(G(z,{\theta _G}),{\theta _D}))]\} \]
Keeping D fixed, update G by\[{\theta _G} \leftarrow {\theta _G} - {\gamma _G}{\nabla _G}L\]where

\[{\nabla _G}L = \frac{\partial }{{\partial {\theta _G}}}{E_{z \sim Pdata(z)}}[log(1 - D(G(z,{\theta _G}),{\theta _D}))]\]

Wasserstein distance is also called the EM(Earth-Mover) distance
% MathType!MTEF!2!1!+-
% feaagKart1ev2aaatCvAUfeBSjuyZL2yd9gzLbvyNv2CaerbuLwBLn
% hiov2DGi1BTfMBaeXatLxBI9gBaerbd9wDYLwzYbItLDharqqtubsr
% 4rNCHbWexLMBbXgBd9gzLbvyNv2CaeHbl7mZLdGeaGqiVu0Je9sqqr
% pepC0xbbL8F4rqqrFfpeea0xe9Lq-Jc9vqaqpepm0xbba9pwe9Q8fs
% 0-yqaqpepae9pg0FirpepeKkFr0xfr-xfr-xb9adbaqaaeGaciGaai
% aabeqaamaabaabauaakeaacaWGxbGaaiikaiaadcfadaWgaaWcbaGa
% amOCaaqabaGccaGGSaGaamiuamaaBaaaleaacaWGNbaabeaakiaacM
% cacqGH9aqpdaWfqaqaaiGacMgacaGGUbGaaiOzaaWcbaGaeq4SdCMa
% eSipIOJaey4dIuTaaiikaiaadchadaWgaaadbaGaamOCaiaacYcaae
% qaaSGaamiCamaaBaaameaacaWGNbaabeaaliaacMcaaeqaaOGaamyr
% amaaBaaaleaacaGGOaGaamiEaiaacYcacaWG5bGaaiykaiablYJi6i
% abeo7aNbqabaGccaGGBbGaeSyjIaLaamiEaiabgkHiTiaadMhacqWI
% LicucaGGDbaaaa!6387!
\[W({P_r},{P_g}) = \mathop {\inf }\limits_{\gamma  \sim \prod ({p_{r,}}{p_g})} {E_{(x,y) \sim \gamma }}[\parallel x - y\parallel ]\]

Where % MathType!MTEF!2!1!+-
% feaagKart1ev2aaatCvAUfeBSjuyZL2yd9gzLbvyNv2CaerbuLwBLn
% hiov2DGi1BTfMBaeXatLxBI9gBaerbd9wDYLwzYbItLDharqqtubsr
% 4rNCHbWexLMBbXgBd9gzLbvyNv2CaeHbl7mZLdGeaGqiVu0Je9sqqr
% pepC0xbbL8F4rqqrFfpeea0xe9Lq-Jc9vqaqpepm0xbba9pwe9Q8fs
% 0-yqaqpepae9pg0FirpepeKkFr0xfr-xfr-xb9adbaqaaeGaciGaai
% aabeqaamaabaabauaakeaadaqeaaqaaiaacIcacaWGqbWaaSbaaSqa
% aiaadkhaaeqaaOGaaiilaiaadcfadaWgaaWcbaGaam4zaaqabaGcca
% GGPaaaleqabeqdcqGHpis1aaaa!46E6!
$\prod {({P_r},{P_g})} $ denotes the set of all joint distributions % MathType!MTEF!2!1!+-
% feaagKart1ev2aaatCvAUfeBSjuyZL2yd9gzLbvyNv2CaerbuLwBLn
% hiov2DGi1BTfMBaeXatLxBI9gBaerbd9wDYLwzYbItLDharqqtubsr
% 4rNCHbWexLMBbXgBd9gzLbvyNv2CaeHbl7mZLdGeaGqiVu0Je9sqqr
% pepC0xbbL8F4rqqrFfpeea0xe9Lq-Jc9vqaqpepm0xbba9pwe9Q8fs
% 0-yqaqpepae9pg0FirpepeKkFr0xfr-xfr-xb9adbaqaaeGaciGaai
% aabeqaamaabaabauaakeaacqaHZoWzcaGGOaGaamiEaiaacYcacaWG
% 5bGaaiykaaaa!44CC!
$\gamma (x,y)$ whose marginal are respectively ${P_r}$ and  ${P_g}$ . Intuitively, $\gamma (x,y)$  indicates how much “mass” must be transported from x to y in order to transform the distributions ${P_r}$   into the distribution ${P_g}$ . The EM distance then is the “cost” of the optimal transport plan.

% MathType!MTEF!2!1!+-
% feaagKart1ev2aaatCvAUfeBSjuyZL2yd9gzLbvyNv2CaerbuLwBLn
% hiov2DGi1BTfMBaeXatLxBI9gBaerbd9wDYLwzYbItLDharqqtubsr
% 4rNCHbWexLMBbXgBd9gzLbvyNv2CaeHbl7mZLdGeaGqiVu0Je9sqqr
% pepC0xbbL8F4rqqrFfpeea0xe9Lq-Jc9vqaqpepm0xbba9pwe9Q8fs
% 0-yqaqpepae9pg0FirpepeKkFr0xfr-xfr-xb9adbaqaaeGaciGaai
% aabeqaamaabaabauaakeaacqGHsislcaWGfbWaaSbaaSqaaiaadIha
% cqWI8iIocaWGWbWaaSbaaWqaaiaadEgaaeqaaaWcbeaakiaacUfaca
% WGMbWaaSbaaSqaaiabeM8a3bqabaGccaGGOaGaamiEaiaacMcacaGG
% Dbaaaa!4C51!

\section{Experiments}

\begin{figure}[ht]
\centering
\includegraphics[width=3.0in]{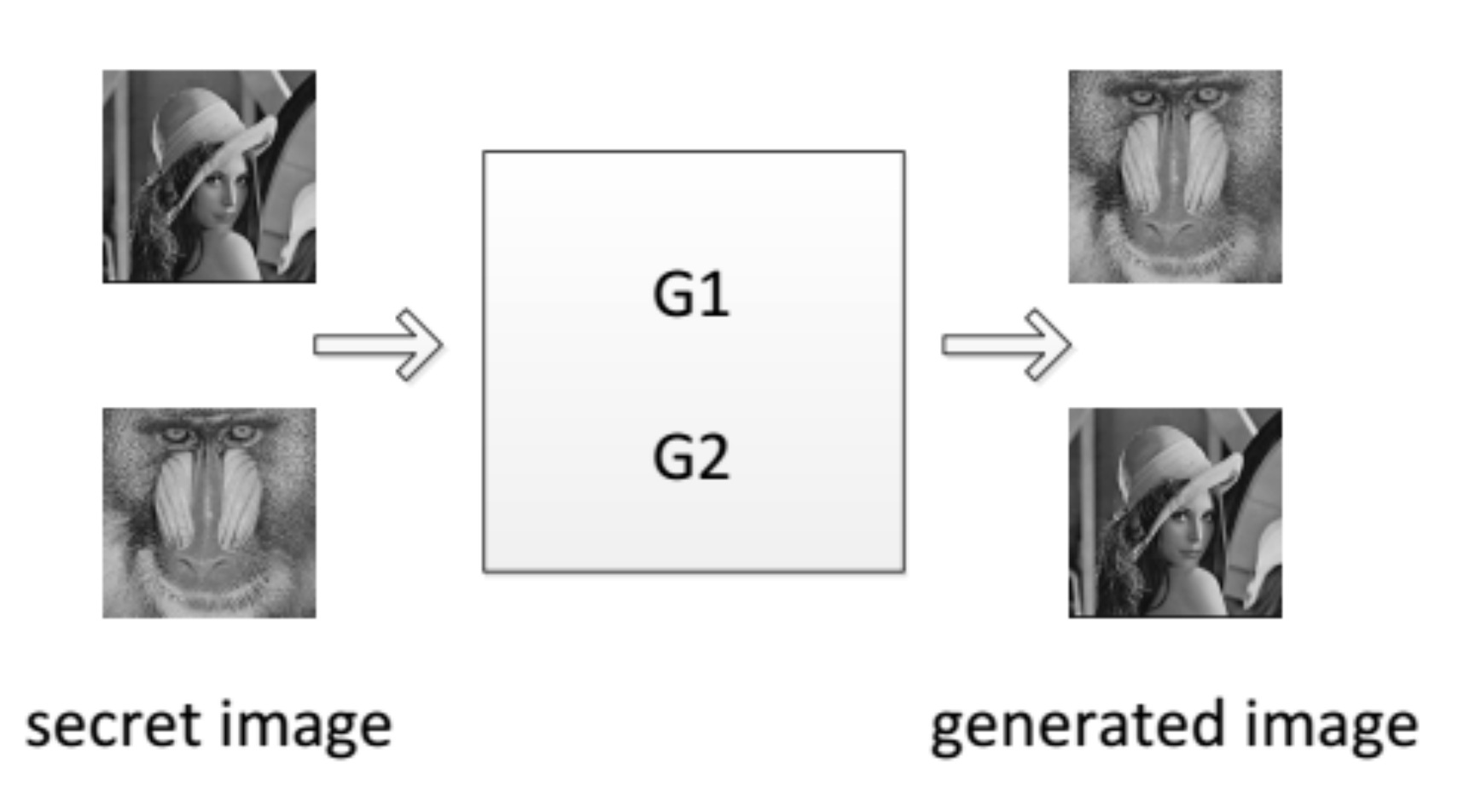}
% where an .eps filename suffix will be assumed under latex,
% and a .pdf suffix will be assumed for pdflatex; or what has been declared
% via \DeclareGraphicsExtensions.
\caption{}
\label{fig_4}
\end{figure}

\begin{figure}[ht]
\centering
\includegraphics[width=3.0in]{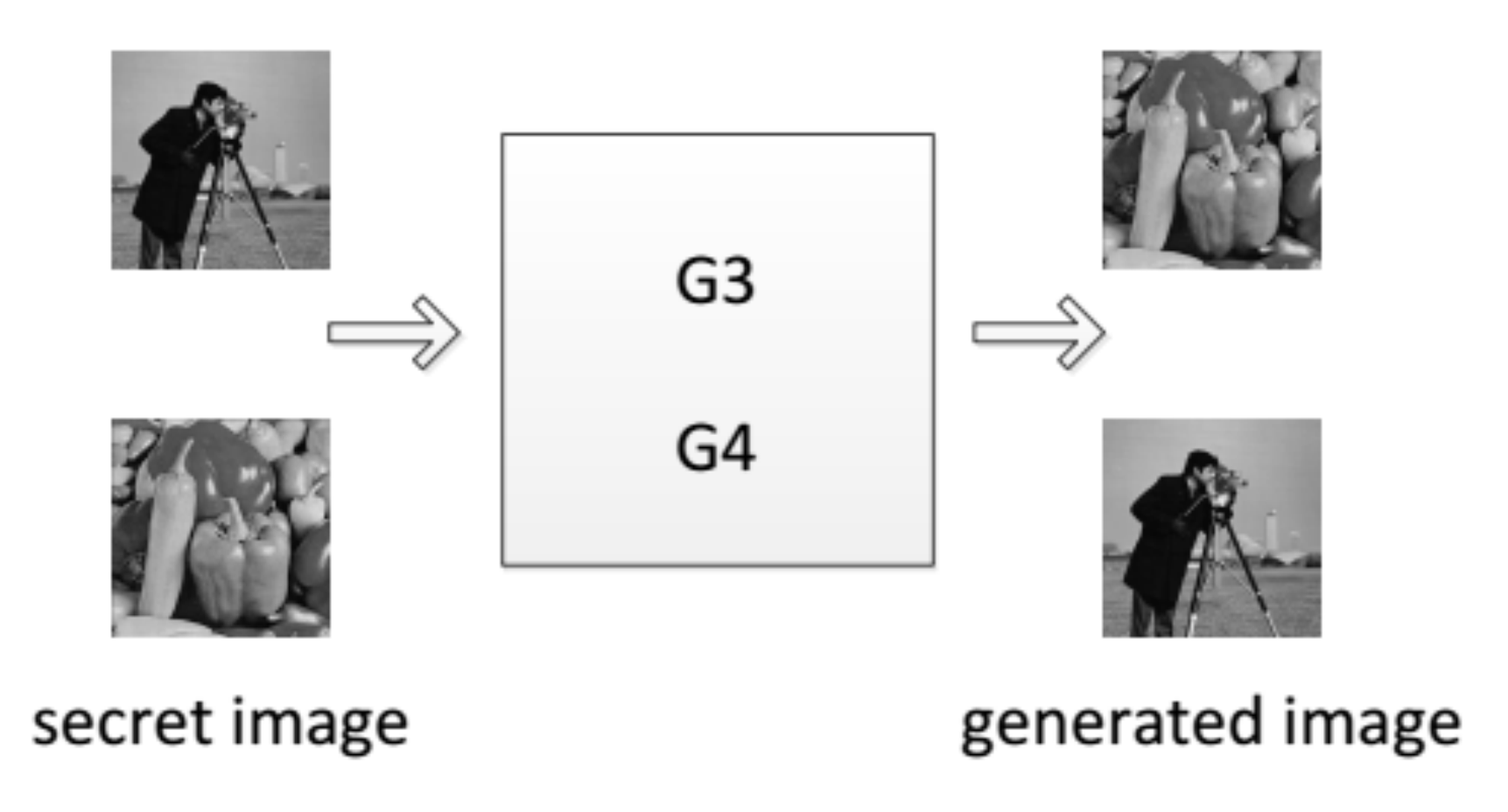}
% where an .eps filename suffix will be assumed under latex,
% and a .pdf suffix will be assumed for pdflatex; or what has been declared
% via \DeclareGraphicsExtensions.
\caption{}
\label{fig_5}
\end{figure}

In this paper, 5,000 images are randomly selected from the CelebA dataset to experiment, and the results show that the coverless image information hiding based on generative model method can be implemented well. The sender and receiver share the same dataset and the same parameters. As shown in Fig.~\ref{fig_4} and Fig.~\ref{fig_5}, we feed the secret image img into the generative model, generating the meaning-normal and independent IMG’ which is not related to the secret image we want to transmit.

\begin{figure}[ht]
\setlength{\abovecaptionskip}{0.cm}
\centering
\includegraphics[width=3.0in]{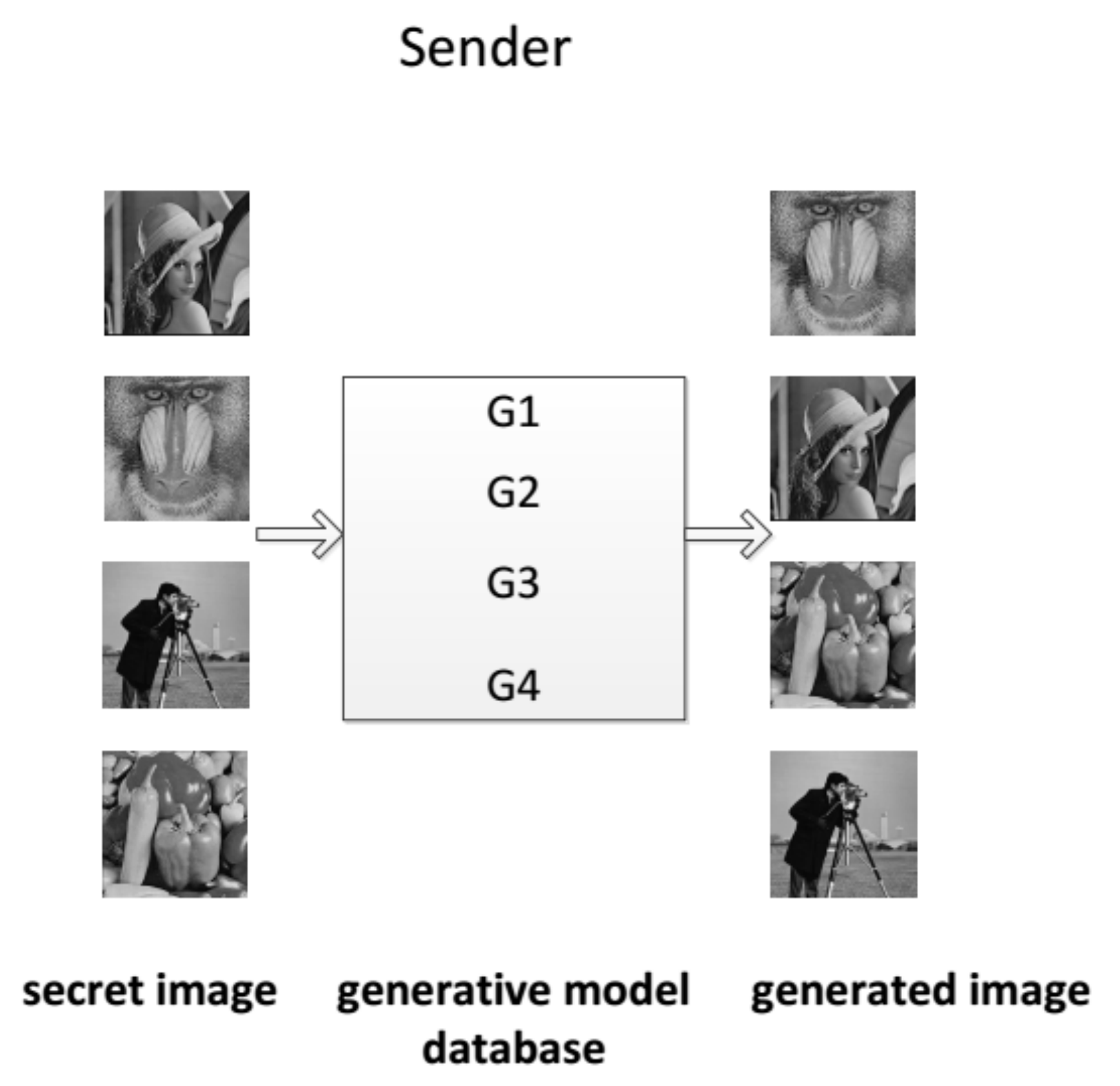}
% where an .eps filename suffix will be assumed under latex,
% and a .pdf suffix will be assumed for pdflatex; or what has been declared
% via \DeclareGraphicsExtensions.
\caption{}
\label{fig_6}
\end{figure}

\begin{figure}[ht]
\setlength{\abovecaptionskip}{0.cm}
\centering
\includegraphics[width=3.0in]{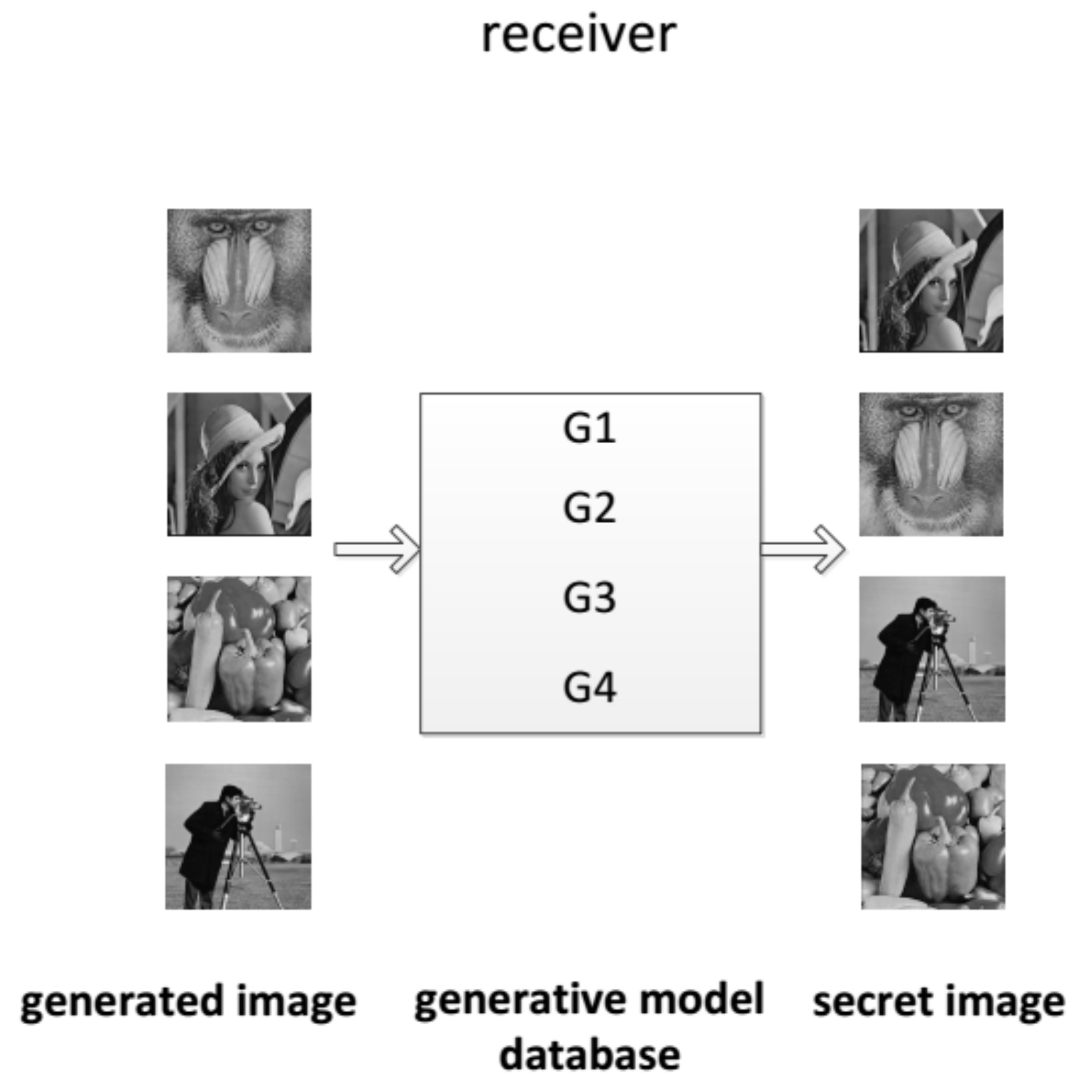}
% where an .eps filename suffix will be assumed under latex,
% and a .pdf suffix will be assumed for pdflatex; or what has been declared
% via \DeclareGraphicsExtensions.
\caption{}
\label{fig_7}
\end{figure}

As shown above, we choose Lena as the secret image img, it can generate the IMG’ visually the same as Baboon we want to transmit. In the meantime, we also trained Baboon to generate the IMG’ visually the same as Lena through the WGAN. We save the corresponding generative model G1 and G2 of generating visually the same as Baboon and Lena respectively. Using the same method, we take the cameraman and peppers as the secret image to experiment respectively, and they can generate corresponding peppers and cameraman. We also save the corresponding generative model G3 and G4 of generating visually the same as peppers and cameraman respectively, and apply them to the next experiment, instead of the WGAN.
We put the generative model G1, G2, G3 and G4 of generating visually the same as Baboon, Lena, peppers and cameraman in a database respectively, so that the generative model database is built. Since both the sender and the receiver train well the generative model database, we perform experiments as shown in Fig.~\ref{fig_6} and Fig.~\ref{fig_7}.

As shown above, when the sender wants to transmit the secret image Lena, the generated image Baboon can be transmitted to the receiver to generate a generated image visually the same as the secret image Lena, similarly, if you want to transmit Baboon, you can transmit the generated image Lena, if you want to transmit cameraman, you can transmit the generated image peppers, if you want to transmit peppers, you can transmit the generated image cameraman.
In this experiment, we have successfully achieved the effect of coverless image information hiding based on generative model method by feeding a secret image to generate a meaning-normal and independent image which is not related to the secret image we want to transmit, and when the secret image is given, the transmitted image is unique and specific. Consequently, the image information hiding method proposed in this paper is feasible. In practical application, we are more concerned with the content of the image rather than the pixels in addition to professional image workers, this method can produce a meaning-normal and independent image which is not related to the secret image we want to transmit, which can satisfy most requirements, thereby, we suppose that if you want to send a secret image, you only need to transmit a meaning-normal and independent image to the receiver, the receiver only need to feed transmitted image to the generative model database, generate an image visually the same as the secret one, no needing direct transmission of the secret image. Besides, the transmitted image does not embed any information of the secret image, so it does not give visual cues to attackers, and the image steganography analysis does not work. This method can resist detection of all the existing steganalysis tools, and improve the security of the image.

\section{Conclusion}

To sum up, the paper proposed the coverless image information hiding based on generative model method. An image visually the same as the secret image is generated by transmitting a normal-meaningful image to the receiver. A fed image corresponds uniquely to a secret image. This method is practical. Therefore, it can be applied to image hiding and image protection.

\section*{Acknowledgment}

This work was supported by the National Natural Science Foundation of China under Grant NO.U1204606, U1404603, the Science and Technology Foundation of Henan Province under Grant No.172102210335.

\ifCLASSOPTIONcaptionsoff
  \newpage
\fi

%\bibliographystyle{IEEEtran}
%\bibliography{ref}


%\end{IEEEbiography}


%\newpage



%\vfill

\end{document}